\documentclass[a4paper]{llncs}
\usepackage{cite}
\usepackage{graphicx}
\usepackage{epstopdf}
\usepackage{upgreek}
\DeclareGraphicsExtensions{.eps}
\usepackage[cmex10]{amsmath}
\usepackage{amsfonts}
\usepackage{amssymb}
\usepackage{enumerate}
\usepackage{url}
\usepackage{multirow}
\usepackage{scrextend}
\usepackage{tabularx}
\usepackage{fixltx2e}
\usepackage{subfigure}

\def\Fig#1{{Fig.\ \ref{fig:#1}}}

\def\Tbl#1{{Table \ref{tbl:#1}}}

\begin{document}

\title{3D Context Enhanced \\Region-based Convolutional Neural Network \\for  End-to-End Lesion Detection}

\author{Ke Yan$ ^1 $ \and Mohammadhadi Bagheri$ ^2 $ \and Ronald M. Summers$ ^1 $}
\institute{$ ^1 $Imaging Biomarkers and Computer-Aided Diagnosis Laboratory \\
	$ ^2 $Clinical Image Processing Service \\
	Department of Radiology and Imaging Sciences \\
	National Institutes of Health Clinical Center, Bethesda, MD 20892-1182
	 \\ \email{\{ke.yan, mohammad.bagheri, rms\}@nih.gov} }

\maketitle
\begin{abstract}
Detecting lesions from computed tomography (CT) scans is an important but difficult problem because non-lesions and true lesions can appear similar. 3D context is known to be helpful in this differentiation task. However, existing end-to-end detection frameworks of convolutional neural networks (CNNs) are mostly designed for 2D images. In this paper, we propose 3D context enhanced region-based CNN (3DCE) to incorporate 3D context information efficiently by aggregating feature maps of 2D images. 3DCE is easy to train and end-to-end in training and inference. A universal lesion detector is developed to detect all kinds of lesions in one algorithm using the DeepLesion dataset. Experimental results on this challenging task prove the effectiveness of 3DCE. We have released the code of 3DCE in \footnote{\url{https://github.com/rsummers11/CADLab/tree/master/lesion_detector_3DCE}}.
\end{abstract}

\section{Introduction}
\label{sec:intro}

Automated lesion detection in computed tomography (CT) scans plays an important role in computer-aided disease screening and tracking. To differentiate lesions from non-lesions, 3D context is crucial \cite{ Ding2017Lung3d, Dou2017multilevel, Liao2017Lung3d, Roth2016RandomView}. However, existing detection frameworks using convolutional neural networks (CNNs) \cite{Ren2015Faster, Dai2016RFCN} are typically designed for 2D images. Therefore, algorithms that can take 3D context information into consideration are in need.

As a direct solution, Liao et al.\ \cite{Liao2017Lung3d} extended the region proposal network (RPN) \cite{Ren2015Faster} to 3D RPN to process volumetric CT data. However, 3D CNNs are very memory-consuming so that sometimes it is hard to fit a single sample into the memory of a mainstream GPU \cite{Liao2017Lung3d}. To solve this problem, \cite{Liao2017Lung3d} used small 3D patches as the input of the network. Besides, 3D bounding-boxes are generally more difficult to annotate than 2D ones, which leads to sparse training data for 3D RPNs. Hence, data augmentation was used in \cite{Liao2017Lung3d} to combat over-fitting. In \cite{Ding2017Lung3d}, 2D networks were first applied to generate lesion candidates, then 3D CNN classifiers were trained for false positive reduction (FPR). Some researchers trained classifiers on the aggregation of multiple 2D slices (e.g.\, three orthogonal views (2.5D) or random views of the candidate lesion) for FPR \cite{Roth2016RandomView}. FPR-based approaches have two stages and are not end-to-end.

In this paper, we propose 3D context enhanced region-based CNNs (3DCE) to incorporate 3D context into 2D regional CNNs. Multiple neighboring slices are sent into a 2D detection network to generate feature maps separately, which are then aggregated for final prediction. We improve the region-based fully convolutional network (R-FCN) \cite{Dai2016RFCN} for this task. 3DCE has many advantages: 1) Compared with the 2-stage candidate generation + FPR approaches, 3DCE is more efficient and end-to-end in both training and inference. 2) It can leverage popular 2D CNN backbones and pretrained weights such as VGG-16 \cite{Simonyan2015VGG}. The weights are learned from millions of images \cite{Deng2009ImgNet} and are known to be beneficial for transfer learning \cite{Shin2016CNN}. On the contrary, 3D CNNs lack such pretrained models and have to be trained from scratch. 3) 3DCE only requires 2D bounding-box annotations to train. Compared with 3D methods, it can obtain large-scale training data more easily, e.g., from radiologists' routine lesion measurements \cite{Yan2018DeepLes}.

Previous studies on lesion detection generally focused on specific types of lesions, such as lung nodules and liver lesions. While some common types receive much attention, many infrequent but still clinically significant types have been ignored. In this paper, we apply the proposed algorithm on DeepLesion \cite{Yan2018DeepLes}, a large-scale and diverse lesion dataset. It contains over 32K 2D annotations of lesions with a variety of types. Using this dataset, we develop a universal lesion detection algorithm that finds all types of lesions with one unified framework. After incorporating 3D contexts with 3DCE, the sensitivity of lesion detection with 4 false positives per image is improved from 80.32\% to 84.37\% on the test set of the challenging dataset, proving the effectiveness of 3DCE.
\section{Method}
\label{sec:method}

Our goal is to consider 3D context in lesion detection, meanwhile leveraging pretrained 2D CNN weights for transfer learning \cite{Shin2016CNN}. A simple solution \cite{Ding2017Lung3d} is to input multiple neighboring slices into object detection frameworks with pretrained backbones. Since current detectors are mostly designed for natural images with three channels (RGB), it is necessary to extend the filters in the first layer and pad it with zeros for the extra input channels. Thus, the new network can start from using the three channels with non-zero weights, and gradually learn the new weights in the first layer to fit the extra input channels. The drawback of this data-level fusion strategy is that the pretrained weights may need to change greatly to adapt to 3D textures. Our idea is to fuse information in the feature map level. We first group slices to 3-channel images and extract good feature maps for each image, then aggregate the feature maps of neighboring images to collect 3D information, and finally build lesion classifiers on top of the fused features. Therefore, the backbone network structure for 2D images can be kept, whereas its representation capability is enhanced for 3D by feature aggregation.

\subsection{3DCE}
\label{subsec:3dce}
\begin{figure}[]
	\begin{center}
		\includegraphics[width=\linewidth,trim=0 0 0 0, clip]{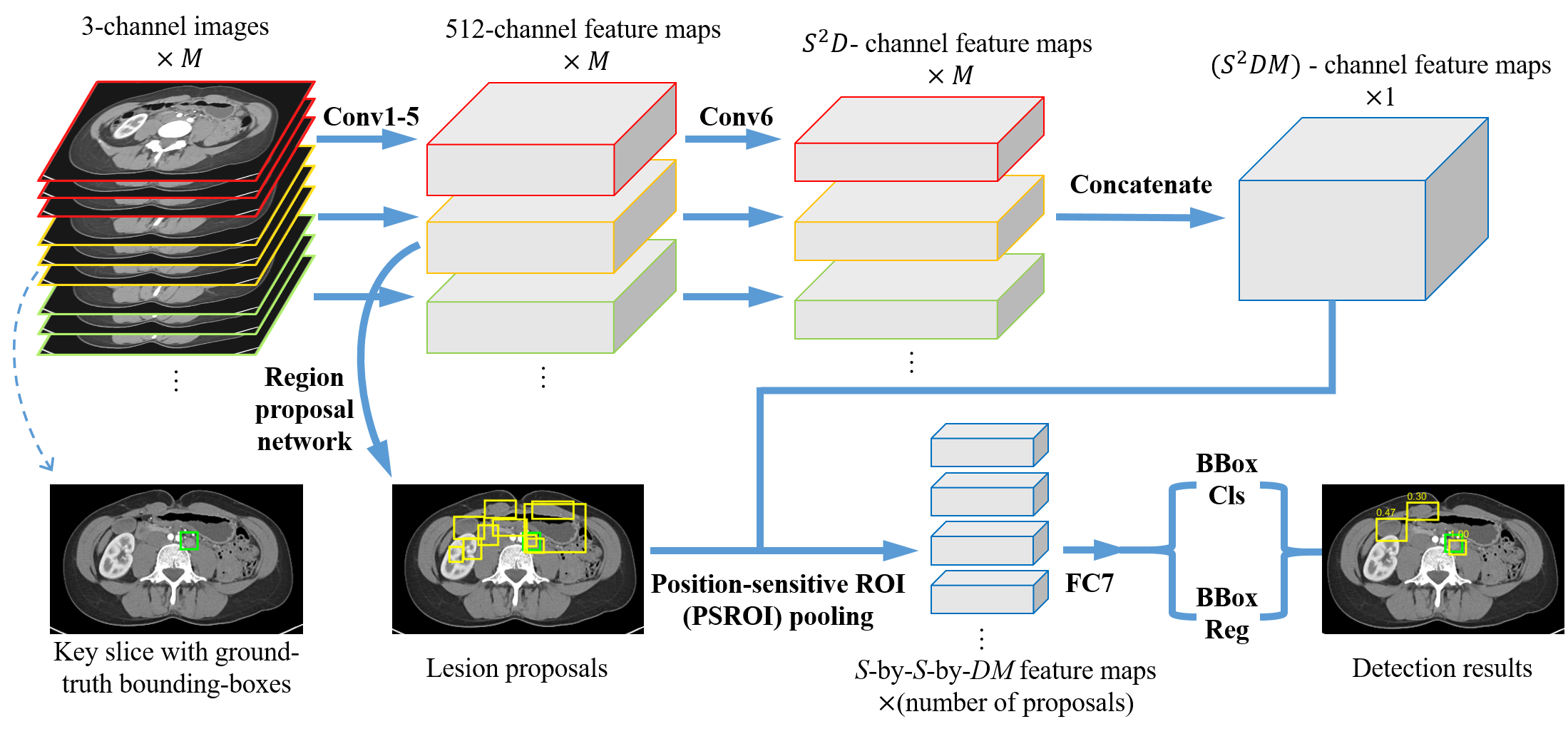} 
	\end{center}
	\caption{The framework of 3DCE for lesion detection.}
	\label{fig:framework}
\end{figure}

The framework of the proposed 3D context enhanced region-based CNN (3DCE) is presented in \Fig{framework}. We adopt the R-FCN \cite{Dai2016RFCN} for this task, which we find is faster, and more accurate and memory-efficient than the widely-used faster region-based CNN (faster RCNN) \cite{Ren2015Faster, Ding2017Lung3d}. Different from faster RCNN, R-FCN constructs a set of position-sensitive score maps, each encoding the object class or position information in a relative spatial position of the object. After that, a position-sensitive region of interest (PSROI) pooling layer summarizes these score maps on each lesion proposal. The object classification and bounding-box regression results are finally obtained by another pooling operation \cite{Dai2016RFCN}.

We first improve R-FCN by adding 3 new layers after PSROI pooling: a 2048D fully-connected (FC) layer, a ReLU layer, and two FC layers for classification and bounding-box regression, respectively. A performance boost (0.7\%) was observed with these additional layers, whereas the speed is comparable to the original R-FCN because we have made the feature maps before FC7 thinner. The last two pooling layers (pool4 and pool5) in VGG-16 are removed to enhance the resolution of the feature map, since lesions are often small and sparse. In 3DCE, $ 3M $ slices are grouped to $ M $ 3-channel images, as shown in \Fig{framework}. During training, the central slice contains the ground-truth bounding-box and the other slices provide the 3D context. We combine the $ M $ images to be a sample to input into the convolutional blocks (Conv1--5 of VGG-16) to produce $ M $ feature maps. Only the feature map derived from the central image is sent to the region proposal network (RPN) to generate lesion proposals. All feature maps undergo another convolutional layer (Conv6). They are then concatenated to aggregate the 3D information, forming an $ S^2DM $-channel feature map, where $ S=7 $ is the size of the pooled feature map for each proposal. $ D $ controls the number of 2D feature maps of 3DCE. We empirically set $ D=10 $ in this paper. $ M $ determines the amount of 3D context to be incorporated. Larger $ M $ brings more information and more memory cost and risk of over-fitting. We found $ M=3\sim 9 $ (18$ \sim $54mm) a reasonable range in our lesion detection task.

In 3DCE, the weights of Conv1--6 are shared for different images in a sample. This strategy reduces the number of parameters to be learned compared with the 3D filters in 3D CNNs. Thus, 3DCE is less prone to overfitting. The 2D feature extractor (Conv1--6) and 3D feature classifier (FC7--end) are trained simultaneously, so Conv1--6 can learn useful features for both lesions and their contexts. This strategy can also be viewed as factorizing 3D filters to 2D+1D \cite{Qiu2017P3D}. There are four loss terms in 3DCE: lesion classification loss and bounding-box regression loss in RPN and improved R-FCN. They are optimized jointly.

\subsection{Implementation Details}
\label{subsec:detail}

The algorithms were implemented using MXNet \cite{Chen2015MXNet} and run on an NVIDIA Titan X Pascal GPU. We initialized the weights in Conv1--Conv5 with an ImageNet \cite{Deng2009ImgNet} pretrained VGG-16 model. All other layers were randomly initialized. Five anchor scales (16, 24, 32, 48, 96) and three anchor ratios (1:2, 1:1, 2:1) were used in RPN. The loss weight of bounding-box regression in improved R-FCN was set to be 10. In training, each mini-batch had 2 samples (each sample had $ M $ 3-channel images) when $ M<7 $ and 1 sample when $ M\geq 7 $. We adopted the stochastic gradient descent (SGD) optimizer and set the base learning rate to 0.001, then reduced it by a factor of 10 after the 4th and 5th epochs.

\section{Experiments}
\label{sec:detail}

\subsection{DeepLesion Dataset}
\label{subsec:ds}

The DeepLesion dataset \cite{Yan2018DeepLes} available at \footnote{\url{https://nihcc.box.com/v/DeepLesion}} was mined from a hospital's picture archiving and communication system (PACS) based on bookmarks, which are markers annotated by radiologists during their daily work to highlight significant image findings. It is a large-scale dataset with 32,735 lesions on 32,120 axial slices from 10,594 CT studies of 4,427 unique patients. Different from existing datasets that typically focus on one type of lesion, DeepLesion contains a variety of lesions including those in lungs, livers, kidneys, etc., and enlarged lymph nodes in the chest, abdomen, and pelvis (see examples in \Fig{det_res}). Their diameters range from 0.21 to 342.5mm. The great diversity in type and size makes lesion detection in this dataset a challenging task. We rescaled the 12-bit CT intensity range to floating-point numbers in [0,255] using a single windowing (-1024--3071 HU) that covers the intensity ranges of the lung, soft tissue, and bone. Every image slice was resized so that each pixel corresponds to 0.8mm. The slice intervals of most CT scans in the dataset are either 1mm or 5mm. We interpolated in the $ z $-axis to make the intervals of all volumes 2mm. The black borders in images were clipped for computation efficiency. We divided DeepLesion into training (70\%), validation (15\%), and test (15\%) sets by randomly splitting the dataset at the patient level. No data augmentation was performed.

\subsection{Results and Discussion}
\label{subsec:res}

\begin{figure}[]
	\begin{center}
		\includegraphics[width=.7\linewidth,trim=0 0 0 29, clip]{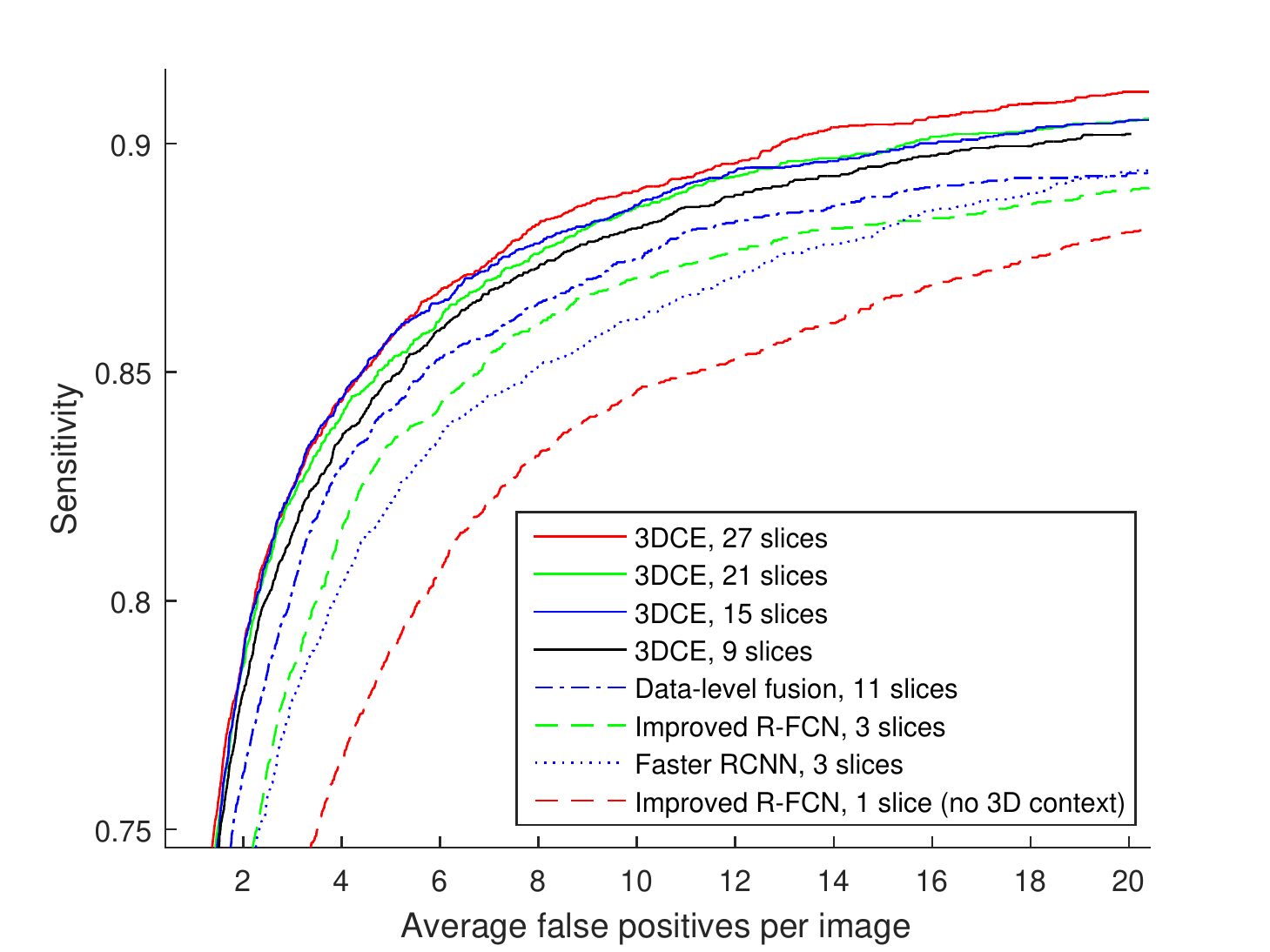} 
	\end{center}
	\caption{FROC curves of various methods on the test set of DeepLesion (4802 images).}
	\label{fig:froc}
\end{figure}

A predicted box was regarded as correct if its intersection over union (IoU) with a ground-truth box is larger than 0.5. The free-response receiver operating characteristic (FROC) curves of several methods are shown in \Fig{froc}.
\begin{itemize}
	\item[$ \bullet $] Improved R-FCN, 1 slice: Only the key slice with the lesion annotation was used for training and inference, thus no 3D context information was exploited. Its inferior accuracy indicates the importance of 3D context.
	\item[$ \bullet $] Faster RCNN, 3 slice: The baseline method. Images composed of 3 neighboring slices were input to faster RCNN \cite{Ren2015Faster} with VGG-16 backbone. Pool4 and pool5 were removed similar to the improved R-FCN.
	\item[$ \bullet $] Improved R-FCN, 3 slices: See Sec.\ \ref{subsec:3dce}. It outperformed faster RCNN using the same 3-slice 3D context.
	\item[$ \bullet $] Data-level fusion, 11 slices: In this method, multiple slices are input into improved R-FCN. We found that 11 slices achieved the best performance, but it is still lower than 3DCE with 9 input slices, proving that the feature fusion strategy of 3DCE can better leverage the 3D context information.
	\item[$ \bullet $] 3DCE: We tested applying 9, 15, 21, 27 slices ($ M=3,5,7,9 $). They achieved the best accuracy among all methods compared on the dataset.
\end{itemize}

To analyze the detection accuracy on different lesions and images, we split the test set according to three criteria and display the results in \Tbl{cls}. Eight lesion types are provided for the test set of DeepLesion. It is found that lung, mediastinum, and liver lesions have high sensitivity, probably because their intensity and appearance is relatively distinctive from the background. The sensitivity of all types were improved by 3DCE. Bone lesions were improved the most. 
As for lesion size, smaller lesions ($ <10 $mm) are harder to detect and benefited more by 3D context. 3DCE works better on CT scans with finer slice intervals because more precise information can be provided by the intermediate slices, compared with the interpolated slices in scans with bigger intervals.

\begin{table}[]
	\centering
	\scriptsize	
	\setlength{\tabcolsep}{2.7pt}
	\renewcommand{\arraystretch}{1.2}
	\caption{Sensitivity (\%) at 4 false positives (FPs) per image on the test set of DeepLesion. The baseline is the faster RCNN algorithm. Lesions were sorted according to their types, sizes, and the slice intervals of the CT scans. The abbreviations of lesion types stand for lung, mediastinum, liver, soft tissue, pelvis, abdomen, kidney, and bone, respectively \cite{Yan2018DeepLes}. The mediastinum type mainly consists of lymph nodes in the chest. Abdomen lesions are miscellaneous ones that are not in liver or kidney. The soft tissue type contains lesions in the muscle, skin, and fat.}
	\begin{tabular}{cccccccccccccccc} 
		
		\hline 
		&	\multicolumn{8}{c}{Lesion type} && \multicolumn{3}{c}{Lesion diameter (mm)} && \multicolumn{2}{c}{Slice interval (mm)} \\
		
		&	LU	&	ME	&	LV	&	ST	&	PV	&	AB	&	KD	&	BN	&&	$<10$	&	$10\sim30$	&	$>30$	&& $ <2.5 $	&	$ >2.5 $	\\
		\hline
		Baseline &	88	&	84	&	80	&	76	&	76	&	75	&	72	&	55	&&	72	&	83	&	80	&&	80	&	80	\\
		3DCE	&	91	&	88	&	84	&	82	&	81	&	80	&	76	&	63	&&	78	&	86	&	84	&&	85	&	83	\\			
		
		\hline
	\end{tabular}

	\label{tbl:cls}
\end{table}

During experiments, we noticed that sometimes the detector identified smaller parts of a large or scattered lesion with a big ground-truth bounding-box. Although the IoU is less than 0.5 in such cases, the detection may still be viewed as a true positive (TP) and it can also help the radiologists. To overcome this evaluation bias, we utilized the intersection over the detected bounding-box area ratio (IoBB) as another criterion. Besides, there are missing lesion annotations in the test set, because DeepLesion is based on radiologists' bookmarks, who typically mark only representative lesions in their daily work. Thus, an FP prediction may actually be a TP. We invited two experienced radiologists to re-annotate all lesions on 300 random slices in the test set of DeepLesion. The number of ground-truth lesions in the 300 slices grew from 305 to 768. We also assessed the algorithms on this all-lesion test set. The results are shown in \Tbl{acc}.

\begin{table}[]
	\centering
	\setlength{\tabcolsep}{5pt}
	\renewcommand{\arraystretch}{1.2}
	\caption{Sensitivity (\%) at 4 FPs per image of different methods. Both the original test set and the all-lesion test set are investigated using two overlap computation criteria.}
	\begin{tabular}{p{4cm}cccccc} 
		\hline 
		&	\multicolumn{2}{c}{Original test set} && \multicolumn{2}{c}{All-lesion test set} & \multirow{1}{1.5cm}{Inference time (ms)} \\
		& IoU	& IoBB	&& IoU	& IoBB	&  \\
		\hline
		No 3D context	&	76.51	&	80.16	&	& 	66.90	&	72.47	&	\bf 19
		\\
		Faster RCNN, 3 slices	&	80.32	&	85.34	&	& 	71.60	&	80.31	&	32
		\\
		Improved R-FCN, 3 slices	&	81.53	&	85.89	&	& 	74.39	&	81.88	&	\bf 19
		\\
		Data-level fusion, 11 slices	&	82.94	&	86.52	&	& 	74.56	&	80.49	&	28 \\
		3DCE, 9 slices	&	83.57	&	87.81	&	& 	\bf 76.31	&	\bf 82.75	&	56 \\
		3DCE, 27 slices	&	\bf 84.37	&	\bf 87.85	&	& 	75.09	&	\bf 82.75	&	114 \\
		\hline
	\end{tabular}

	\label{tbl:acc} 
\end{table}

In \Tbl{acc}, 3DCE still has the best accuracy despite the test sets and evaluation criteria. The performance computed using IoBB is better than IoU. The sensitivity at 4 FPs on the all-lesion test set actually became lower than the original test set of DeepLesion. DeepLesion consists of only representative lesions that radiologists think are measurable in their daily work, which are often subjective choices. Meanwhile, the all-lesion set was intensively labeled to include every abnormality. Some lesions in the all-lesion set are not measurable or too small, thus do not exist in the training set, which affects the algorithms' performance. In other words, both the number of FPs and the sensitivity are decreased by the new annotations. The inference time is also shown in \Tbl{acc}. The improved R-FCN spent significantly less time than the baseline faster RCNN (since faster RCNN needs to run 4 FC layers on a thick feature map) while still got better accuracy. 3DCE's time complexity is roughly linearly proportional to the number of input slices, since it generates feature maps for multiple images. However, if tested on volumetric data, this extra time cost can be largely reduced because the feature maps of neighboring slices can be cached and reused.

2D candidate generation + 3D/2.5D false positive reduction (FPR) methods were also tested. The accuracy of the FPR classifiers on this dataset is not promising, which is possibly due to the small inter-class variance (lesions and non-lesions look very similar) and large intra-class variance (many lesion types) of the candidates. We also designed a 3D CNN that receives 27-slice inputs (same as 3DCE), extracts features using 3D filters, and predicts 2D boxes on the key slice. It was adapted from improved R-FCN and trained from scratch. Its sensitivity at 4 FPs per image is 79.7\% compared with 3DCE's 84.4\%, proving that 3DCE with pretrained weights and factorized filters is superior.

\begin{figure}[]
	\centering
	\subfigure[]{\includegraphics[width = 0.28\textwidth, trim=0 45 0 10mm, clip]{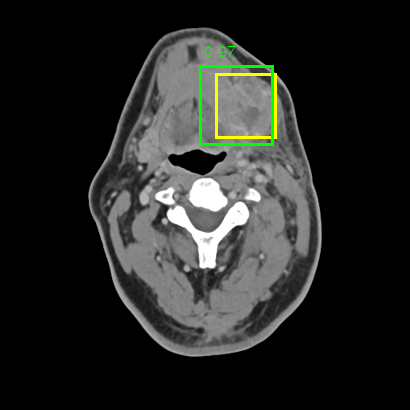}}
	\subfigure[]{\includegraphics[width = 0.28\textwidth, trim=80 5 0 135, clip]{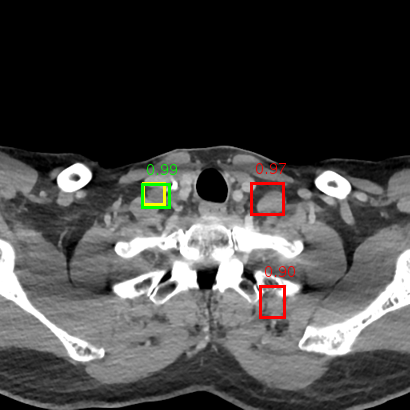}}
	\subfigure[]{\includegraphics[width = 0.28\textwidth, trim=80 100 50 80, clip]{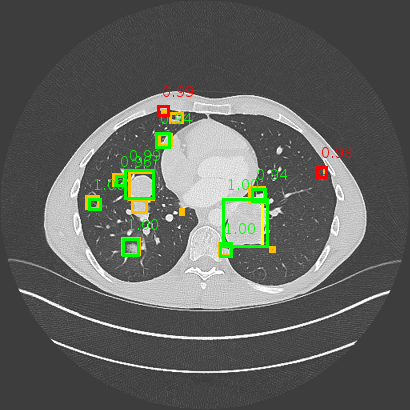}}
	\subfigure[]{\includegraphics[width = 0.28\textwidth, trim=60 60 50 105, clip]{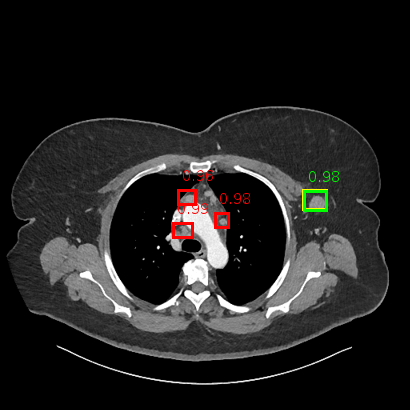}} 
	\subfigure[]{\includegraphics[width = 0.28\textwidth, trim=30 75 30 50, clip]{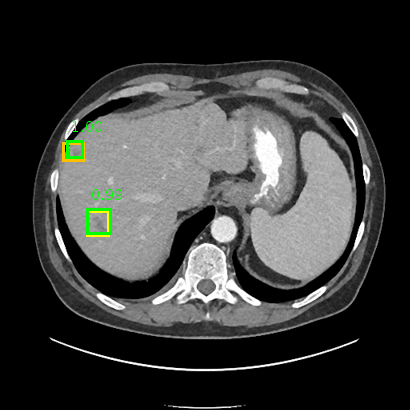}}
	\subfigure[]{\includegraphics[width = 0.28\textwidth, trim=80 80 40 80, clip]{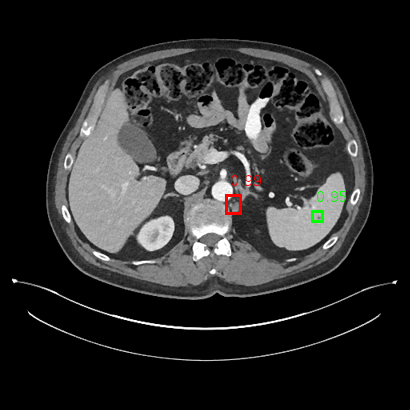}}
	\subfigure[]{\includegraphics[width = 0.28\textwidth, trim=30 75 30 50, clip]{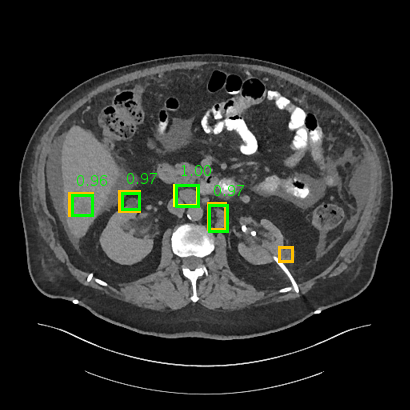}} 
	\subfigure[]{\includegraphics[width = 0.28\textwidth, trim=60 60 60 100, clip]{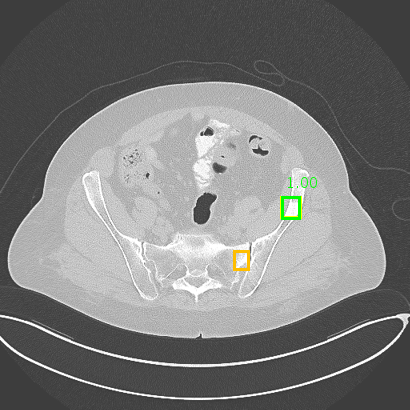}} 
	\subfigure[]{\includegraphics[width = 0.28\textwidth, trim=60 60 60 100, clip]{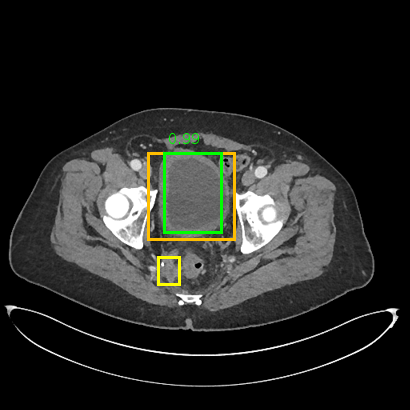}} 
	\caption{Detection results of 3DCE with scores $ >0.9 $ on the test set of DeepLesion. Yellow, orange, green, and red boxes correspond to ground-truths in the test set, additional ground-truths in the all-lesion test set, predicted true positives, and false positives.
	}
	\label{fig:det_res} 
\end{figure}

Sample detection results are shown in \Fig{det_res}. There are a large variety of lesions in DeepLesion. In (c), 14 lung nodules/masses were annotated and 10 were detected (the two FPs are actually TPs with bigger boxes). In (d), an axillary enlarged lymph node was detected but three small mediastinum ones were incorrectly spotted. The detector sometimes cannot distinguish enlarged and normal lymph nodes due to its scale robustness.

\subsection{Results on the official data split of DeepLesion}
\label{subsec:official_res}

\begin{figure}[]
	\begin{center}
		\includegraphics[width=.7\linewidth,trim=0 0 30 30, clip]{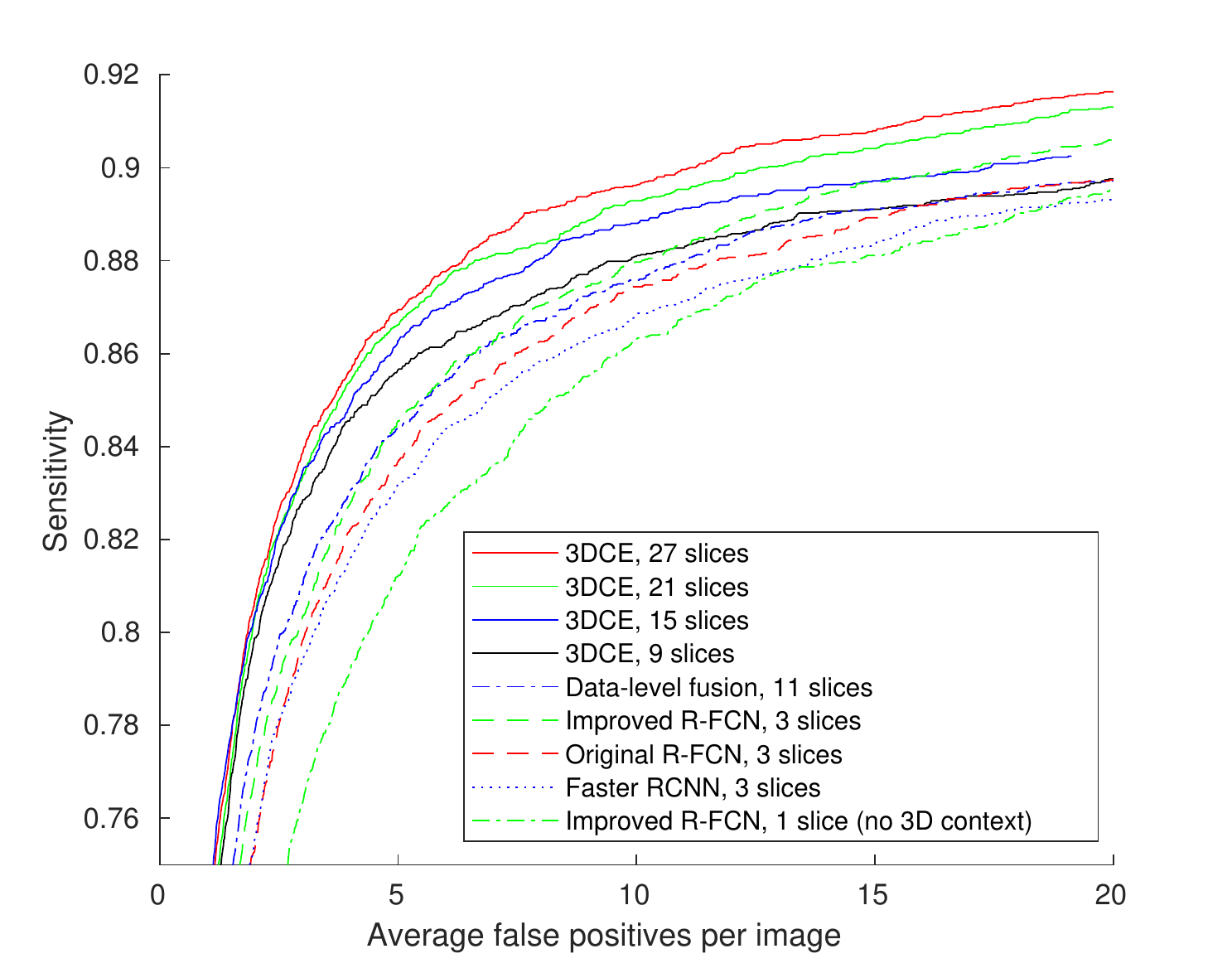} 
	\end{center}
	\caption{FROC curves of various methods on the official test set of DeepLesion (4817 images).}
	\label{fig:froc_official}
\end{figure}

After the release of the DeepLesion dataset and its official random patient-level data split \footnote{\url{https://nihcc.box.com/v/DeepLesion}}, we ran lesion detection experiments again and append the results in this section. We removed the 35 noisy lesion annotations mentioned in the dataset. After that, there are 4,912 lesions from 4,817 images in the test set. Experimental configurations were kept the same as the last section. The original R-FCN was also compared. Results are displayed in \Fig{froc_official}, \Tbl{acc_official}, and \Tbl{cls_official}. It can be found that the general trends of the results remain similar, except that the slice interval is no longer a critical factor for 3DCE. The changes in the results may be due to the difference in data split, the removal of noisy samples, and random factors.

\begin{table}[]
	\centering
	\setlength{\tabcolsep}{5pt}
	\renewcommand{\arraystretch}{1.2}
	\caption{Sensitivity (\%) at various FPs per image on the test set of the official data split of DeepLesion. IoU was used as the overlap computation criterion.}
	\begin{tabular}{p{4cm}cccccc}
		\hline
		FPs per image	& 0.5	& 1		& 2		& 4		& 8		& 16 \\
		\hline
		No 3D context	& 48.60	& 60.57	& 71.19	& 79.15	& 84.77	& 88.42	\\
		Faster RCNN, 3 slices	& 56.90	& 67.26	& 75.57	& 81.62	& 85.83	& 88.74	\\
		Original R-FCN, 3 slices	& 55.70	& 67.26	& 75.37	& 82.21	& 86.26	& 89.19	\\
		Improved R-FCN, 3 slices	& 56.49	& 67.65	& 76.89	& 82.76	& 87.03	& 89.82	\\
		Data-level fusion, 11 slices	& 58.49	& 70.03	& 77.89	& 83.02	& 86.71	& 89.19	\\
		3DCE, 9 slices	& 59.32	& 70.68	& 79.09	& 84.34	& 87.81	& 89.62	\\
		3DCE, 27 slices	& \bf 62.48	& \bf 73.37	& \bf 80.70	& \bf 85.65	& \bf 89.09	& \bf 91.06	\\
		\hline
	\end{tabular}

\label{tbl:acc_official}
\end{table}

\begin{table}[]
	\centering
	\scriptsize	
	\setlength{\tabcolsep}{2.7pt}
	\renewcommand{\arraystretch}{1.2}
	\caption{Sensitivity (\%) at 4 false positives per image on the official test set of DeepLesion. The baseline is the faster RCNN algorithm. 
	}
	\begin{tabular}{cccccccccccccccc} 
		
		\hline 
		&	\multicolumn{8}{c}{Lesion type} && \multicolumn{3}{c}{Lesion diameter (mm)} && \multicolumn{2}{c}{Slice interval (mm)} \\
		
		& LU & ME & LV & ST & PV & AB & KD & BN && $<10$ & $10\sim30$ & $>30$ && $ <2.5 $ & $ >2.5 $ \\
		\hline
		Baseline & 86 & 83 & 88 & 70 & 80 & 79 & 79 & 65 && 75 & 84 & 81 && 81 & 82 \\
		3DCE     & 89 & 88 & 90 & 74 & 84 & 84 & 82 & 75 && 80 & 87 & 84 && 86 & 86 \\
		
		\hline
	\end{tabular}
	
	\label{tbl:cls_official}
\end{table}

\section{Conclusion}
\label{sec:conclusion}
In this paper, we presented 3D context enhanced region-based CNN (3DCE) to leverage the 3D context when detecting lesions in volumetric data. 3DCE is memory-friendly, end-to-end, and simple to implement and train. It consistently improved the detection accuracy on the DeepLesion dataset. We expect it to be applicable in various detection problems where 3D context is helpful. We also developed a detector that can help radiologists find all types of lesions with one unified framework. It may serve as an initial screening tool and send its detection results to other specialist systems trained on certain types of lesions.

{\small \textbf{Acknowledgments:} This research was supported by the Intramural Research Program of the NIH Clinical Center. We thank NVIDIA for the GPU card donation.}
\bibliographystyle{splncs03}
\bibliography{3DCE}
\end{document}